\def\paperlanguage{}
\pdfoutput=1 

\documentclass[letterpaper, 10 pt, conference]{ieeeconf}  

\usepackage{bm}
\usepackage{cite}
\usepackage{flushend}
\include{preamble}

\newcommand{\ctext}[1]{\raise0.2ex\hbox{\textcircled{\scriptsize{#1}}}}

\IEEEoverridecommandlockouts                              

\title{\LARGE \textbf
  {
    \switchlanguage%
    {%
      QDD-MM: A Simple and Compact Open-Source Muscle Module\\with Quasi-Direct-Drive Motor for Tendon-driven Robots
    }%
    {%
      腱駆動ロボットのための準ダイレクトドライブモータを用いた\\シンプルでコンパクトなオープンソース筋モジュール
    }%
  }
}

\author{Kento Kawaharazuka$^{1}$, Yoshimoto Ribayashi$^{1}$, Takahiro Hattori$^{1}$, Shunnosuke Yoshimura$^{1}$, and Kei Okada$^{1}$
  \thanks{$^{1}$ The authors are with the Department of Mechano-Informatics, Graduate School of Information Science and Technology, The University of Tokyo, 7-3-1 Hongo, Bunkyo-ku, Tokyo, 113-8656, Japan.
    {\texttt\small [kawaharazuka, ribayashi, t-hattori, yoshimura, k-okada]@jsk.imi.i.u-tokyo.ac.jp}
  }
}

\begin{document}

\maketitle
\thispagestyle{empty}
\pagestyle{empty}

\begin{abstract}
\switchlanguage%
{%
  Tendon-driven robots, which actuate like biological muscles by using motors to wind wires, offer advantages such as variable stiffness and redundant actuation.
  Despite these advantages, musculoskeletal structures inherently involve a large number of diverse muscles, making it desirable to compactly modularize each muscle unit and flexibly accommodate diverse muscle arrangements.
  While various small muscle modules have been proposed to date, existing designs tend to be complex, difficult to maintain, and suffer from low backdrivability and limited actuation speed due to their high gear ratios.
  In this work, we propose QDD-MM, a muscle module that leverages recent advances in quasi-direct-drive motors to achieve a simpler and more compact structure with higher backdrivability and faster actuation than previous modules.
  By utilizing quasi-direct-drive motors, the module eliminates the need for dedicated force-sensing mechanisms, significantly simplifying the overall design, while also supporting flexible wire routing in multiple directions, various module-to-module connections, and even the use of the module body itself as a structural component.
  QDD-MM integrates these capabilities into a reusable muscle module, and its design is openly available at \href{https://github.com/haraduka/qddmm}{\textcolor{magenta}{github.com/haraduka/qddmm}}.
  We developed tendon-driven robots with one joint and either two or three muscles, as well as a partial musculoskeletal humanoid, using the proposed QDD-MM and verified their effectiveness.
  We expect that this muscle module will democratize tendon-driven robotics and further accelerate progress in the field.
}%
{%
  モータによりワイヤを巻き取ることで筋肉のように駆動する腱駆動ロボットは, 可変剛性や冗長駆動などの利点を有する.
  一方で, 筋骨格構造は多数の多様な筋肉を扱うため, 各筋肉をコンパクトにモジュール化し, 多様な筋配置に対応できることが望ましく, これまで様々な小型の筋モジュールが提案されてきた.
  しかし, これまでの筋モジュールは複雑でメンテナンス性に欠け, ギア比が高いためバックドライバビリティが低く, 動作速度も遅いという課題があった.
  そこで本研究では, 現在の準ダイレクトドライブモータの発展を活用した, これまでよりシンプルかつコンパクトで, 高いバックドライバビリティと高速動作を実現する筋モジュールQDD-MMを提案する.
  準ダイレクトドライブモータを用いることで筋張力測定を不要として構造を大幅に簡素化し, かつ多様な方向へのワイヤ引き出し, 多様なモジュール接続, モジュール自体の構造としての利用も実現している.
  QDD-MMはこれらの機能を再利用可能な筋モジュールとして統合し, オープンソース化により, 誰もが利用できるようにしている.
  このQDD-MMを用いた1関節2筋または1関節3筋の腱駆動ロボット, さらには筋骨格ヒューマノイドの一部を制作し, その有効性を確認した.
  この筋モジュールにより, 今後腱駆動ロボットがより民主化され, さらなる発展が期待される.
}%
\end{abstract}

\section{INTRODUCTION}\label{sec:introduction}
\switchlanguage%
{%
  A variety of tendon-driven humanoids that closely mimic human anatomical structures have been developed to date \cite{gravato2010ecce1, jantsch2013anthrob, asano2016kengoro, kawaharazuka2019musashi}.
  Beyond purely biological imitation, many tendon-driven robots have been created to leverage the advantages of wire-driven actuation \cite{hirose1991tendon, kobayashi1998tendon, endo2019superdragon, temma2024saqiel, sinoue2024cubix}.
  These robots offer numerous benefits, such as enabling variable stiffness control through nonlinear elastic elements \cite{jacobsen1990control, koganezawa1999stiffness}, maintaining functional operation even after muscle failure due to redundancy \cite{kawaharazuka2022redundancy}, allowing flexible changes in body morphology by reconfiguring wire routing \cite{sinoue2024cubix}, and reducing the weight of distal body segments by proximally locating actuators \cite{endo2019superdragon, temma2024saqiel}.

  However, musculoskeletal structures tend to become complex because they involve a large number of diverse muscles.
  Thus, it becomes crucial to compactly modularize each muscle and to ensure that the muscle module can accommodate a wide variety of possible muscle arrangements.
  If the same muscle module can be used across different body parts, the design and fabrication of tendon-driven robots become significantly easier, and maintainability is improved.

  Motivated by these challenges, several small muscle modules have been proposed in previous work \cite{ohta2010improvement, jantsch2013anthrob, asano2015sensordriver, kawaharazuka2017forearm}.
  These modules commonly have adopted a slender cylindrical form factor and integrate a BLDC motor, a custom motor driver, a pulley for wire winding, and a tension sensor.
  Although relatively compact and capable of supporting diverse wiring configurations, these modules employ small motors paired with high gear-ratio transmissions for miniaturization.
  As a result, they suffer from low backdrivability and limited actuation speed.
  Furthermore, the high gear ratio makes it difficult to estimate muscle tension from motor current alone, necessitating dedicated force sensors.
  This increases structural complexity and reduces maintainability.
  QDD actuation has also been applied to cable-driven robots.
  Mi et al. developed a modular tensegrity robot with QDD cable actuators, enabling force estimation and variable stiffness control without external force sensors \cite{mi2025design}.
  Sawaguchi et al. developed Vlimb, a wire-driven wearable robot using AK60-6 actuators and reconfigurable wire routing to balance lifting capability and reachability \cite{sawaguchi2024vlimb}.
  These studies establish the applicability of QDD actuation to cable-driven systems.
  Our focus is an open-source, reusable muscle module with configurable wire outlets and module connections, together with a direct experimental comparison against a conventional high-gear-ratio muscle module.
}%
{%
  これまで, 人体構造を詳細に模倣した, 様々な腱駆動ヒューマノイドが開発されてきた\cite{gravato2010ecce1, jantsch2013anthrob, asano2016kengoro, kawaharazuka2019musashi}.
  加えて, 生物摸倣に限らず, そのワイヤ駆動構造の利点を活かした様々な腱駆動ロボットも開発されてきている\cite{hirose1991tendon, kobayashi1998tendon, endo2019superdragon, temma2024saqiel, sinoue2024cubix}.
  これらのロボットは非線形弾性要素を活用した可変剛性制御が可能であったり\cite{jacobsen1990control, koganezawa1999stiffness}, 筋肉の冗長性によって筋破断時にも継続して動作が可能であったり\cite{kawaharazuka2022redundancy}, ワイヤ配置を変更することで自由に身体形態を変えられたり\cite{sinoue2024cubix}, アクチュエータを近位に配置することで身体を軽量化したり\cite{endo2019superdragon, temma2024saqiel}等, 様々な利点を有している.

  一方で, 特に筋骨格構造は多数の多様な筋肉を扱うため, その身体構造は複雑になりがちである.
  そのため, 各筋肉をどれだけコンパクトにモジュール化できるか, そしてその筋モジュールがどれだけ多様な筋配置に対応できるかが重要となる.
  同じ筋モジュールを様々な身体部位に活用できれば, 腱駆動ロボットの設計や製作が容易になり, メンテナンス性も向上する.

  このような問題意識から, これまでいくつかの小型の筋モジュールが提案されてきた\cite{ohta2010improvement, jantsch2013anthrob, asano2015sensordriver, kawaharazuka2017forearm}.
  これらはどれも, 細長い筒状の形状をしており, BLDCモータと自作のモータドライバ, ワイヤを巻き取るプーリ, 張力センサを搭載している.
  これらは比較的小型で, かつ様々な筋配置に対応できるようになっている.
  しかし, 小型化のために小さなモータを大きなギア比の減速機と一緒に駆動しているため, バックドライバビリティが低く, 動作速度も遅いという課題があった.
  また, ギア比が高いことで, 電流のみから筋張力を推定することが難しく, 筋張力センサを搭載せざるを得ないため, 構造が複雑でメンテナンス性に欠けるという課題もあった.
  一方, QDDをケーブル駆動ロボットに適用した先行研究も存在する.
  Miらは, 外部力センサを用いずに力推定と可変剛性制御を行うQDDケーブルアクチュエータを備えたモジュール型テンセグリティロボットを開発した\cite{mi2025design}.
  Sawaguchiらは, AK60-6アクチュエータと変更可能なワイヤ経路により持ち上げ能力と到達範囲を両立するワイヤ駆動装着型ロボットVlimbを開発した\cite{sawaguchi2024vlimb}.
  これらに対し本研究では, ワイヤ引き出し方向とモジュール接続を変更できる再利用可能な筋モジュールのオープンソース設計と, 従来の高減速比筋モジュールとの直接的な実験比較に焦点を当てる.
}%

\begin{figure}[t]
  \centering
  \includegraphics[width=0.95\columnwidth]{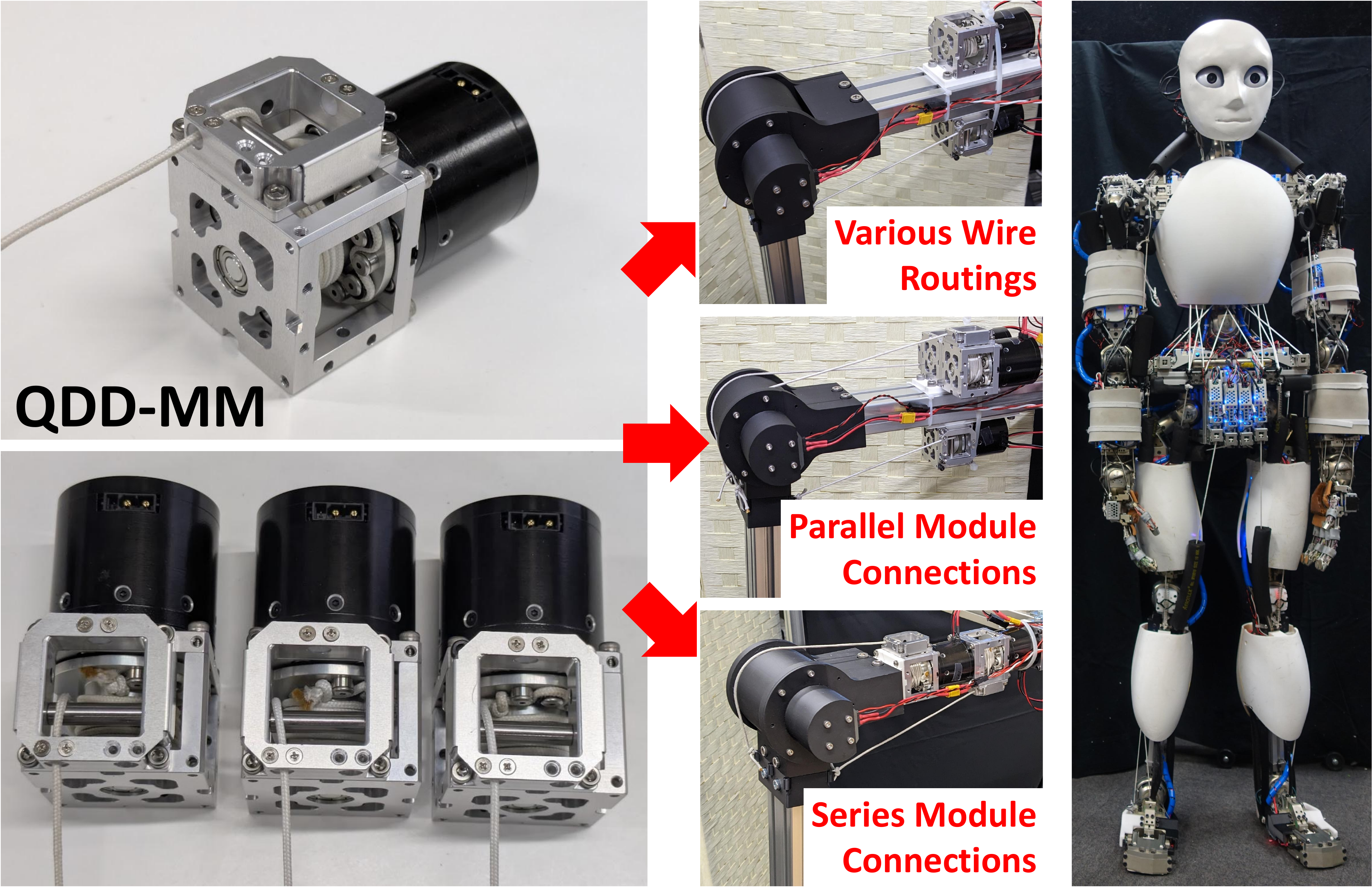}
  \vspace{-1.0ex}
  \caption{The concept of QDD-MM: a compact muscle module (left) and configurable tendon-driven mechanisms (middle). The full humanoid (right) illustrates the intended application.}
  \label{figure:concept}
  \vspace{-3.0ex}
\end{figure}

\begin{figure*}[t]
  \centering
  \includegraphics[width=1.9\columnwidth]{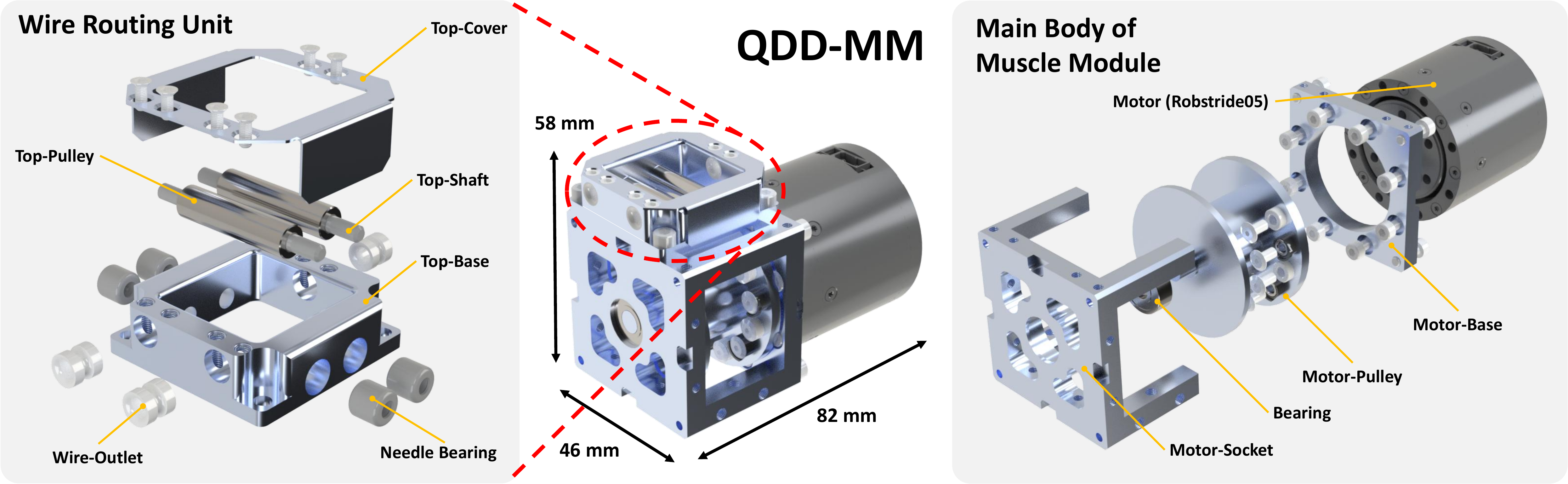}
  \vspace{-1.0ex}
  \caption{Design overview of QDD-MM: the muscle module consists of a main body integrating a quasi-direct-drive motor and pulley system, and a detachable wire routing unit that defines the wire's pulling direction.}
  \label{figure:design-overview}
  \vspace{-3.0ex}
\end{figure*}

\switchlanguage%
{%
  In this work, we propose Quasi-Direct-Drive Muscle Module (QDD-MM), a muscle module that leverages quasi-direct-drive motors \cite{katz2019minicheetah} to achieve a simpler and more compact design, while providing high backdrivability and fast actuation that exceed those of previous muscle modules.
  Recent advances in quasi-direct-drive motors have significantly improved torque density and efficiency, making it possible to generate sufficient muscle force even with low gear ratios.
  Consequently, by reducing the gear ratio, backdrivability can be improved and dedicated force-sensing mechanisms can be removed, which greatly simplifies the structure of the muscle module.
  In addition, QDD-MM retains the capabilities of previous modules, such as flexible wire routing in multiple directions and support for various module-to-module connections.
  It also introduces a mechanism that allows the module itself to serve as a structural component of the robot body.
  The contribution of QDD-MM lies in integrating configurable wire routing, parallel and series module connections, and structural use of the module body into a compact, open-source design available through the Supplementary Materials.
  The high torque density and backdrivability originate primarily from the commercial QDD actuator; our module design makes these properties usable across different tendon-driven structures through common mechanical interfaces.
  \figref{figure:concept} illustrates the concept of constructing diverse tendon-driven robots from a common muscle module, with the full humanoid on the right representing a future application.
  Using QDD-MM, we constructed tendon-driven robots with one joint and either two or three muscles, as well as a partial musculoskeletal humanoid, and we verified their effectiveness experimentally.
  We show that a simple one-joint--two-muscle tendon-driven robot can be extended by using an additional muscle through modular connections, and that a compact tendon-driven robot can be realized by using the muscle module itself as a skeletal structure.
  These results suggest that tendon-driven robots, which have long been considered difficult to work with, may become increasingly accessible and continue to advance through the use of the proposed module.
}%
{%
  そこで本研究では, 準ダイレクトドライブモータを活用した, これまでよりシンプルかつコンパクトで, 高いバックドライバビリティと高速動作を実現する筋モジュールQuasi-Direct-Drive Muscle Module, QDD-MMを提案する.
  近年は準ダイレクトドライブモータの高トルク密度化と高効率化が進んでおり, 低い減速比でも十分な筋張力を発揮できると考えた.
  それに伴い, 減速比を下げることでバックドライバビリティを向上させ, 筋張力センサを不要として筋モジュールの構造を大幅に簡素化することが可能である.
  さらに, これまでの筋モジュールにも備わっている, 多様な方向へのワイヤ引き出し, 多様なモジュール接続の機能を残しつつ, モジュール自体をロボットの構造として活用できるような仕組みも実現している.
  本QDD-MMの貢献は, 多様なワイヤ引き出し方向, 並列・直列のモジュール接続, モジュール自体の構造利用をコンパクトなオープンソース設計に統合した点にある.
  高トルク密度と高いバックドライバビリティは主に市販QDDアクチュエータに由来し, 本モジュール設計は共通の機械インタフェースによってこれらの特性を多様な腱駆動構造で利用可能にする.
  \figref{figure:concept}に, 共通の筋モジュールから多様な腱駆動ロボットを構築する本研究の構想を示す. 右側の全身ヒューマノイドは将来的な応用先を表している.
  本研究では, このQDD-MMを用いた1関節2筋または1関節3筋の腱駆動ロボット, さらには筋骨格ヒューマノイドの一部を制作し, その有効性を確認している.
  単純な1関節2筋の腱駆動ロボットに対して, 追加の筋肉をモジュール間接続により増やしたり, 筋モジュール自体を骨格として利用することで, コンパクトに腱駆動ロボットを実現したりできることを示す.
  これまで誰もが非常にとっつきにくいとされてきた腱駆動ロボットが, 本研究により今後民主化され, さらなる発展を遂げることを期待する.
}%

\section{QDD-MM: Simple and Compact Muscle Module with Quasi-Direct-Drive Motor} \label{sec:proposed}

\subsection{Basic Structure of Muscle Module} \label{subsec:basic}
\begin{figure}[t]
  \centering
  \includegraphics[width=0.8\columnwidth]{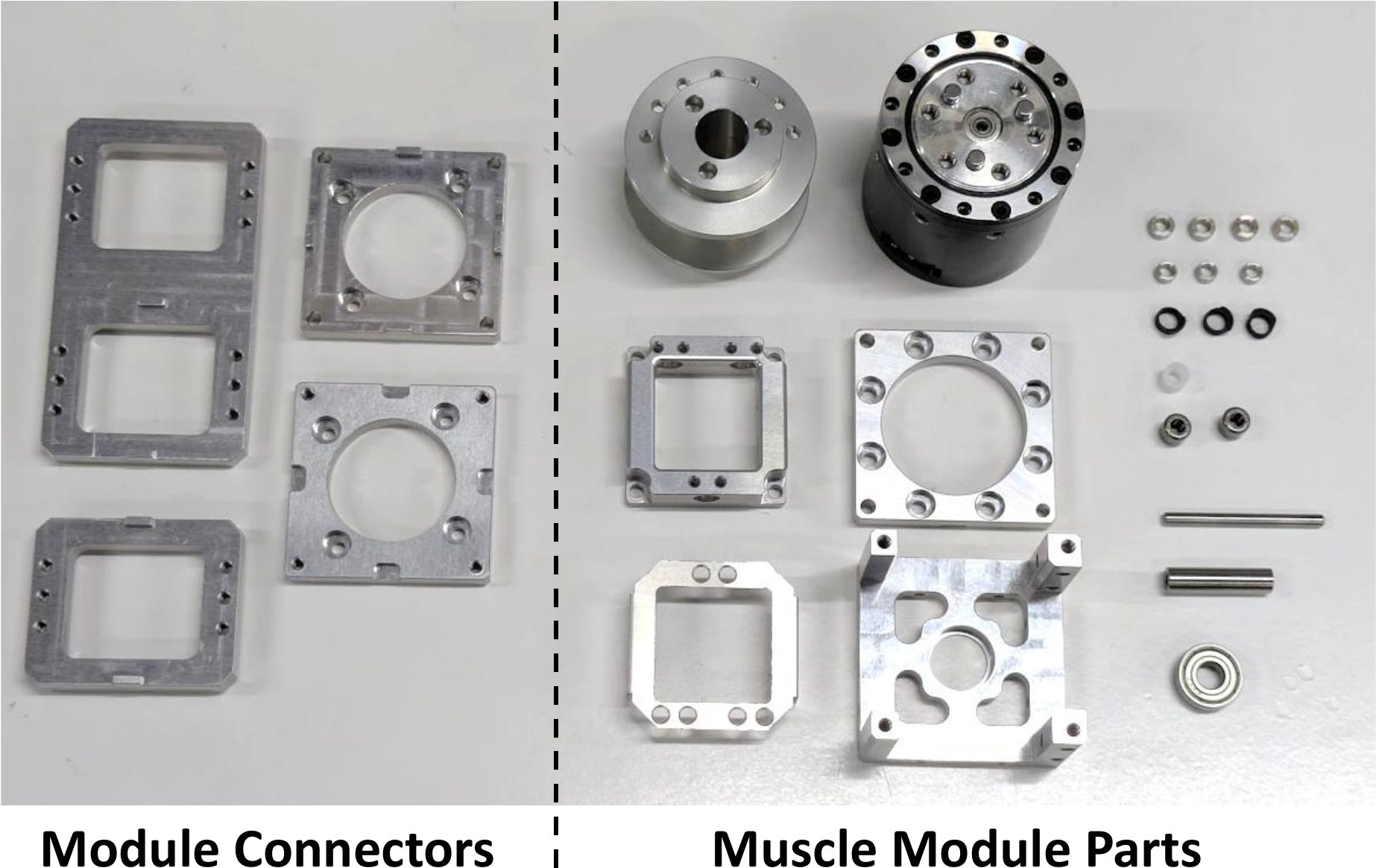}
  \vspace{-1.0ex}
  \caption{All components of QDD-MM consisting of muscle module parts, which constitute the muscle module, and module connectors, which link these modules together.}
  \label{figure:all-parts}
  \vspace{-1.0ex}
\end{figure}

\begin{figure}[t]
  \centering
  \includegraphics[width=0.8\columnwidth]{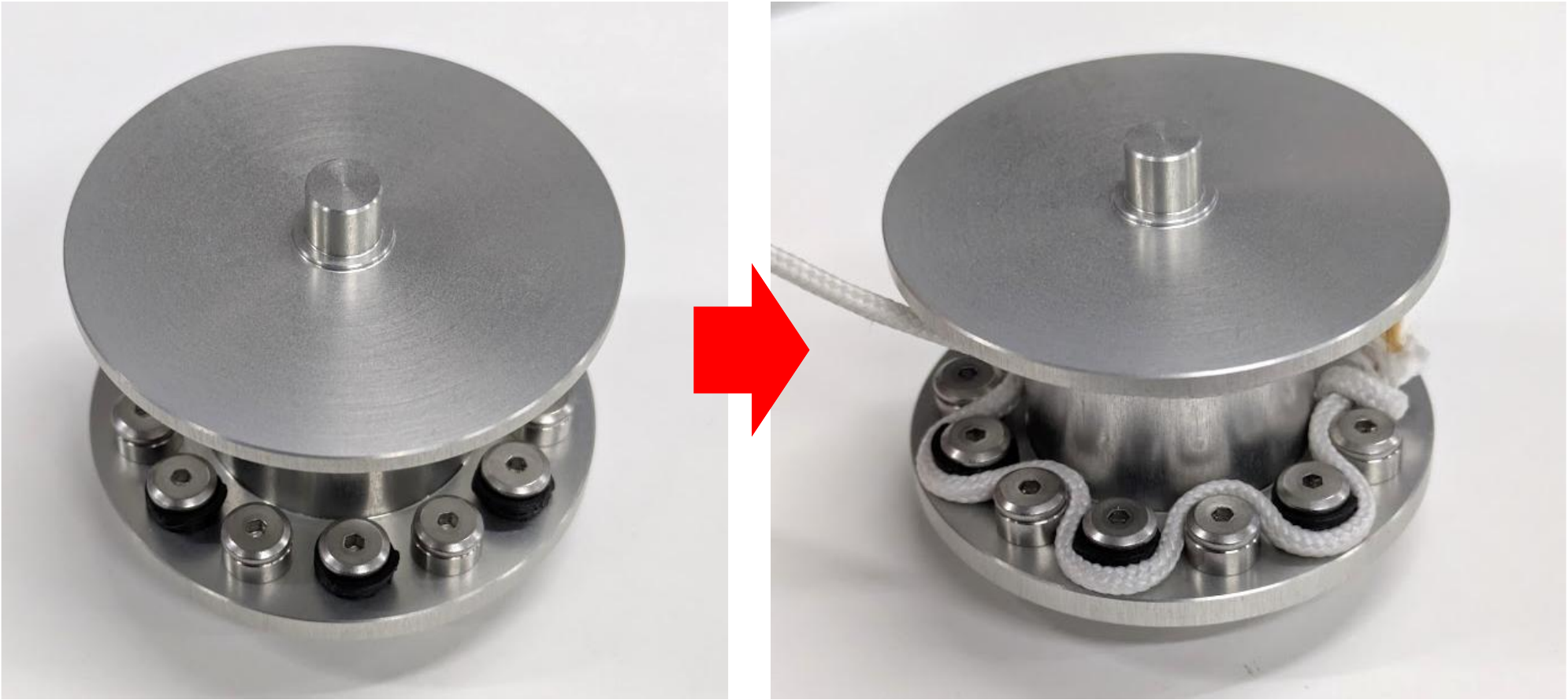}
  \vspace{-1.0ex}
  \caption{Wire connection method to the Motor-Pulley: the wire end is tied, and then passed alternately through seven small pulleys in a weaving manner, allowing the wire to be secured by friction.}
  \label{figure:wire-connection}
  \vspace{-3.0ex}
\end{figure}

\switchlanguage%
{%
  The basic structure of the muscle module QDD-MM developed in this study is shown in \figref{figure:design-overview}, and all components of QDD-MM and its surrounding parts are shown in \figref{figure:all-parts}.
  QDD-MM has a simple and compact design and consists of two major elements: the main body, which includes a quasi-direct-drive motor and the Motor-Pulley for wire winding, and the wire routing unit.
  For the main body, we use a RobStride05 motor.
  The motor driver is integrated into the unit, allowing control via CAN communication.
  A pulley with a diameter of 24 mm is mounted on the motor shaft (the diameter can be changed), and the motor generates muscle tension by winding a 2 mm-diameter Zylon wire around the pulley.
  The motor provides a continuous torque of 1.6 Nm, and with a pulley radius of 12 mm, the maximum continuous tension is 133 N.
  The wire routing unit includes two Top-Pulleys that change the direction of wire extraction, but only one Top-Pulley is needed depending on the configuration.
  The wire can be drawn from any of the three Wire-Outlets.
  To withstand the applied wire tension, small needle bearings are used.
  QDD-MM consists of four machined metal parts and one sheet-metal part, and the module connector used to link multiple QDD-MM units (described in detail in \secref{subsec:connection}) consists of four machined parts.
  All machined components are made of A7075 aluminum alloy, and the sheet-metal components are made of A5052.

  \figref{figure:wire-connection} shows how the wire is attached to the Motor-Pulley.
  The end of the wire is tied and then passed alternately through seven small pulleys, which secure the wire through friction.
  Based on Euler's belt theory, $f_1 = f_0 e^{\mu \theta}$, even when the friction coefficient $\mu=0.3$ between the small pulleys and the wire, seven pulleys provide a total wrap angle of 3.5 turns ($\theta = 3.5 \times 2\pi$).
  As a result, the tension at the wire end $f_0$ becomes $1/731$ of the input tension $f_1$.
  Since the maximum input tension $f_1$ is 133 N, the resulting tension at the wire end $f_0$ is only 0.18 N, which is extremely small.
}%
{%
  本研究で開発した筋モジュールQDD-MMの基本構造を\figref{figure:design-overview}に, QDD-MMとその周辺部品の全てを\figref{figure:all-parts}に示す.
  QDD-MMは非常にシンプルかつコンパクトな構造をしており, 準ダイレクトドライブモータとワイヤを巻き取るMotor-Pulleyから構成されるmain body, wire routing unitの2つの主要な要素から構成されている.
  main bodyについては, モータはRobStride05を用いており, モータドライバは内蔵されていて, CAN通信で制御できる.
  その先端には直径24mmのプーリが取り付けられており(直径は変更することができる), これによりワイヤ(直径2mmのzylon)を巻き取ることで筋張力を発生させる.
  なお, モータの最大連続トルクは1.6Nmであり, プーリ半径が12mmの場合, 最大連続張力は133Nとなる.
  wire routing unitについては, ワイヤを引き出す方向を変えるTop-Pulleyが2本配置でき(使わない場合は1本で良い), 3箇所あるWire-Outletのいずれかからワイヤを引き出すことができる.
  ワイヤ張力に耐えるため小型のニードルベアリングを使用している.
  QDD-MMには4つの切削部品と, 一つの板金部品が使われており, QDD-MM間の接続を行うmodule connector (\secref{subsec:connection}で詳しく説明する)には4種類の切削部品が使われている.
  全ての部品について, 金属切削部品はA7075, 板金部品はA5052を使用している.

  \figref{figure:wire-connection}にMotor-Pulleyへのワイヤの取り付け方を示す.
  ワイヤの端を結び, それを7つの小さなプーリに交互に通すことで, 摩擦によってワイヤを固定している.
  オイラーのベルト理論$f_1=f_0 e^{\mu\theta}$に基づくと, 小さなプーリとワイヤとの間の摩擦係数$\mu$を0.3としても, 7つのプーリにより3.5周の巻き付け($\theta=3.5 \times 2\pi$)が得られ, 末端にかかる張力$f_0$は入力張力$f_1$731分の1になる.
  最大の入力張力$f_1$は133Nであるため, ワイヤ末端にかかる張力$f_0$は0.18Nと非常に小さい.
}%

\subsection{Various Wire Routings} \label{subsec:routing}
\begin{figure}[t]
  \centering
  \includegraphics[width=0.9\columnwidth]{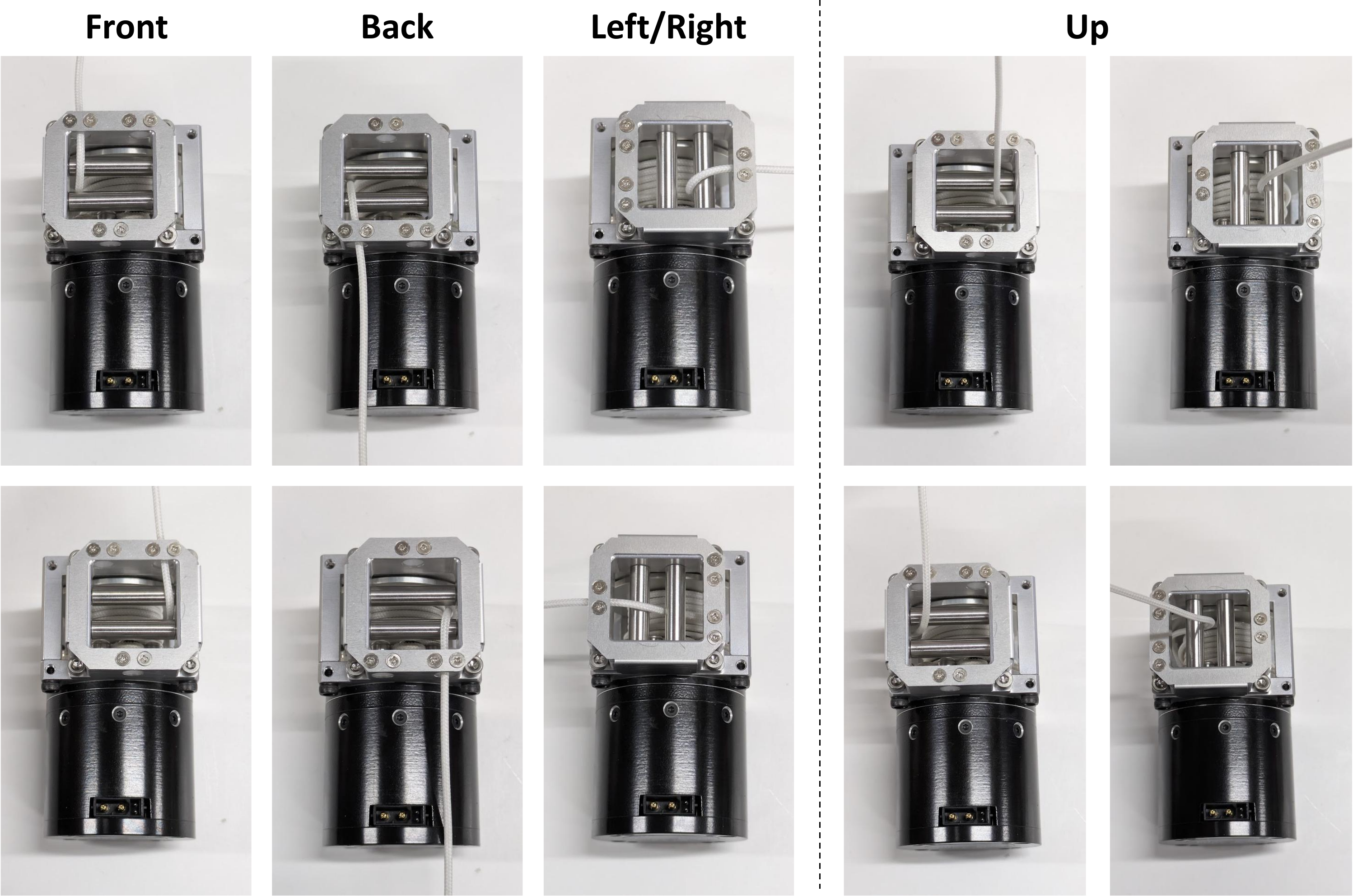}
  \vspace{-1.0ex}
  \caption{Various wire routing directions: by changing the orientation of the wire routing unit, the wire can be routed forward, backward, leftward, rightward, or upward.}
  \label{figure:various-routing}
  \vspace{-3.0ex}
\end{figure}

\switchlanguage%
{%
  QDD-MM allows wires to be routed in various directions through its wire routing unit.
  This capability is highly important, as it significantly increases the design flexibility of tendon-driven robots.
  \figref{figure:various-routing} illustrates several examples of wire extraction directions.
  The wire routing unit can be mounted on either the left or right side, and it can be attached to the main body in four different orientations.
  As a result, the wire can be drawn forward, backward, leftward, or rightward.
  Here, only one of the Top-Pulleys is used.
  When two Top-Pulleys are installed, the wire can also be routed upward instead of being extracted from a Wire-Outlet.
}%
{%
  QDD-MMは, wire routing unitによって様々な方向にワイヤを引き出すことができる.
  この特性は非常に重要であり, 腱駆動ロボットの設計自由度を大きく向上させる.
  \figref{figure:various-routing}に, 様々なワイヤ引き出し方向の例を示す.
  wire routing unitの位置は左右の2種類, main bodyに対する取り付け方向は4方向選べるようになっている.
  これにより, ワイヤを前方, 後方, 左方, 右方に引き出すことができる.
  この際, Top-Pulleyは片方しか使わない.
  また, Top-Pulleyを2つ取りつける場合は, Wire-Outletからではなく, ワイヤを上方に引き出すこともできる.
}%

\subsection{Various Module Connections} \label{subsec:connection}
\begin{figure}[t]
  \centering
  \includegraphics[width=0.75\columnwidth]{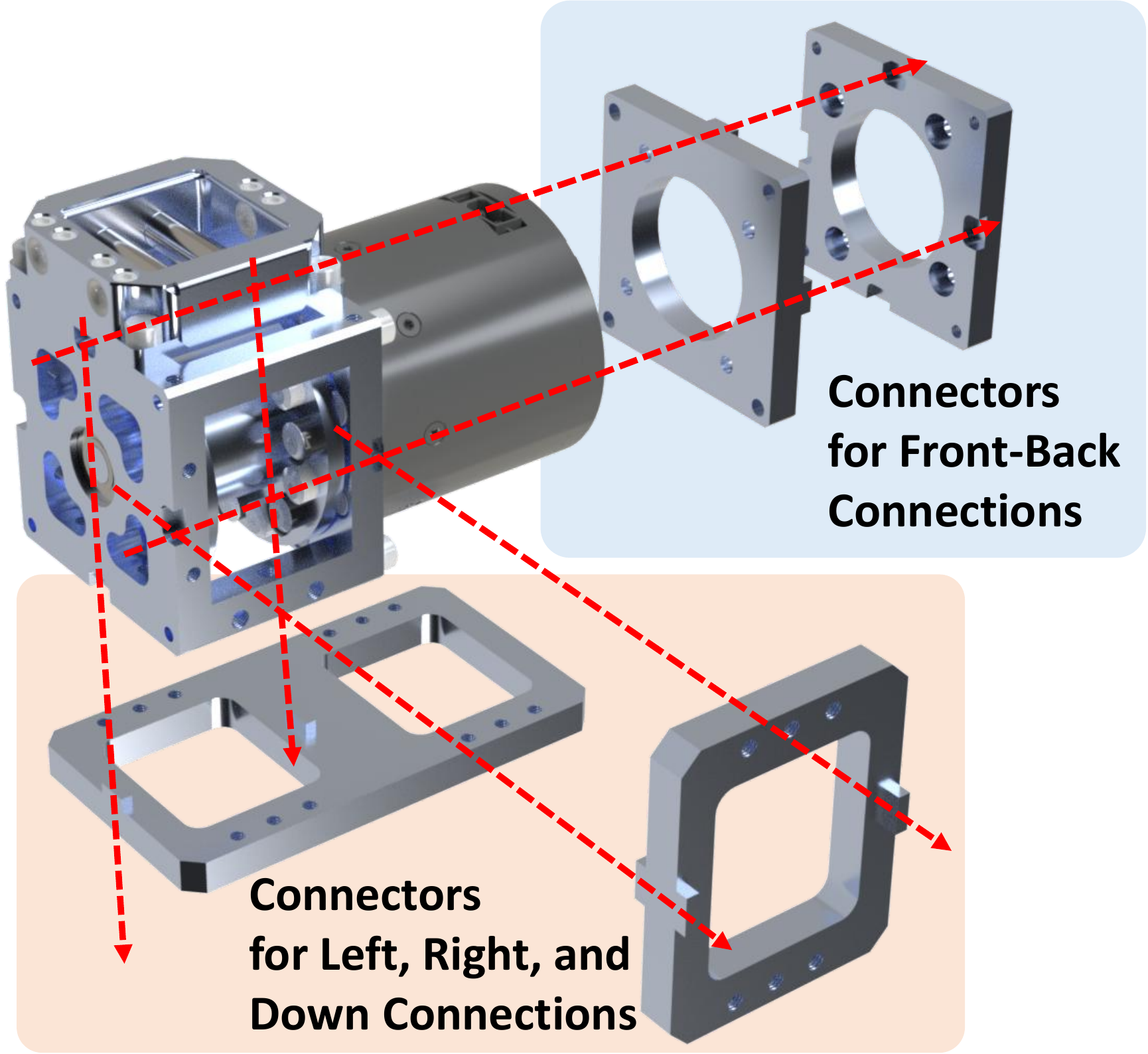}
  \vspace{-1.0ex}
  \caption{The four module connectors have interlocking convex and concave structures, enabling connection between modules in five directions: forward, backward, leftward, rightward, and downward.}
  \label{figure:module-connection}
  \vspace{-3.0ex}
\end{figure}

\begin{figure}[t]
  \centering
  \includegraphics[width=0.8\columnwidth]{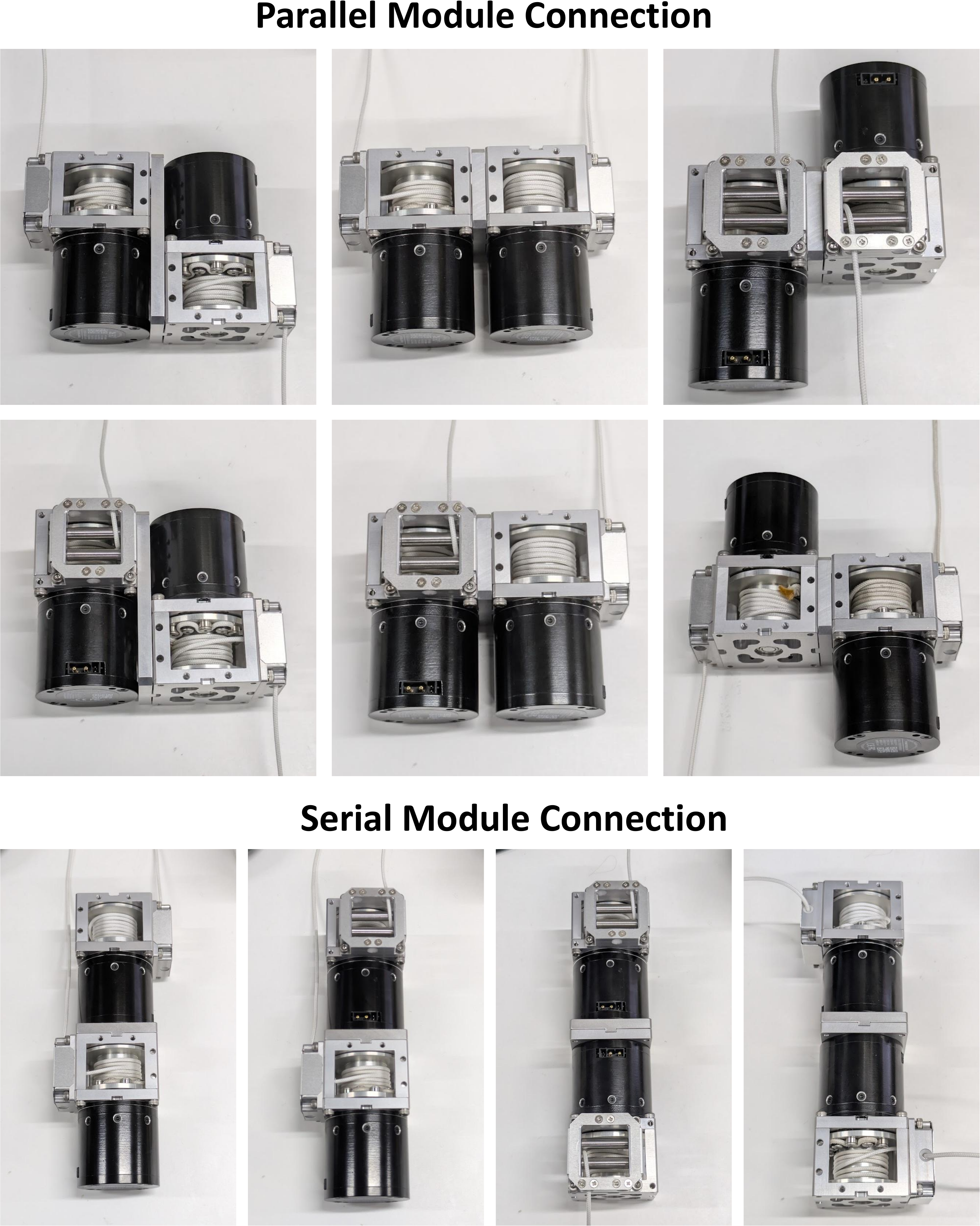}
  \vspace{-1.0ex}
  \caption{Various connections between muscle modules: two muscle modules can be connected to each other to the right, left, and downward using two types of module connectors, and to the front and back using the other two types of module connectors.}
  \label{figure:various-connection}
  \vspace{-3.0ex}
\end{figure}

\switchlanguage%
{%
  QDD-MM units can be freely connected to one another.
  As shown in \figref{figure:module-connection}, the four module connectors have interlocking convex and concave structures that allow two muscle modules to be joined securely with screws.
  Examples of these connections are shown in \figref{figure:various-connection}.
  By using two types of module connectors for the right, left, and downward directions, two muscle modules can be connected in parallel.
  For the forward and backward directions, two additional types of module connectors are used to connect modules in series.
  When combined with different mounting positions for the wire routing unit, this design accommodates a wide range of muscle arrangements.
  The forward and backward connections are particularly important, as they allow the muscle module itself to serve as part of the robot's skeletal structure, thereby enabling seamless integration of bone and muscle components in a compact form.
}%
{%
  QDD-MMは, それら同士を自由に接続することができる.
  \figref{figure:module-connection}に, モジュール間の接続を行うmodule connectorの構造を示す.
  QDD-MM側には凹部があり, 凸部を持つmodule connectorをはめ込み, ネジで止めることで, 2つの筋モジュールを接続できる.
  その接続例を\figref{figure:various-connection}に示す.
  右, 左, 下方には2種類のmodule connectorを用いることで, 2つの筋モジュールを並列に接続できる.
  また, 前方, 後方には別の2種類のmodule connectorを用いることで, 2つの筋モジュールを直列に接続できる.
  wire routing unitの取り付け位置と組み合わせることで, 非常に多くの筋配置に対応することができる.
  特に前方後方の接続は, 筋モジュール自体を骨格として利用することを可能とするため, コンパクトに骨格と筋肉を一体化できる点で重要である.
}%

\begin{figure}[t]
  \centering
  \includegraphics[width=0.8\columnwidth]{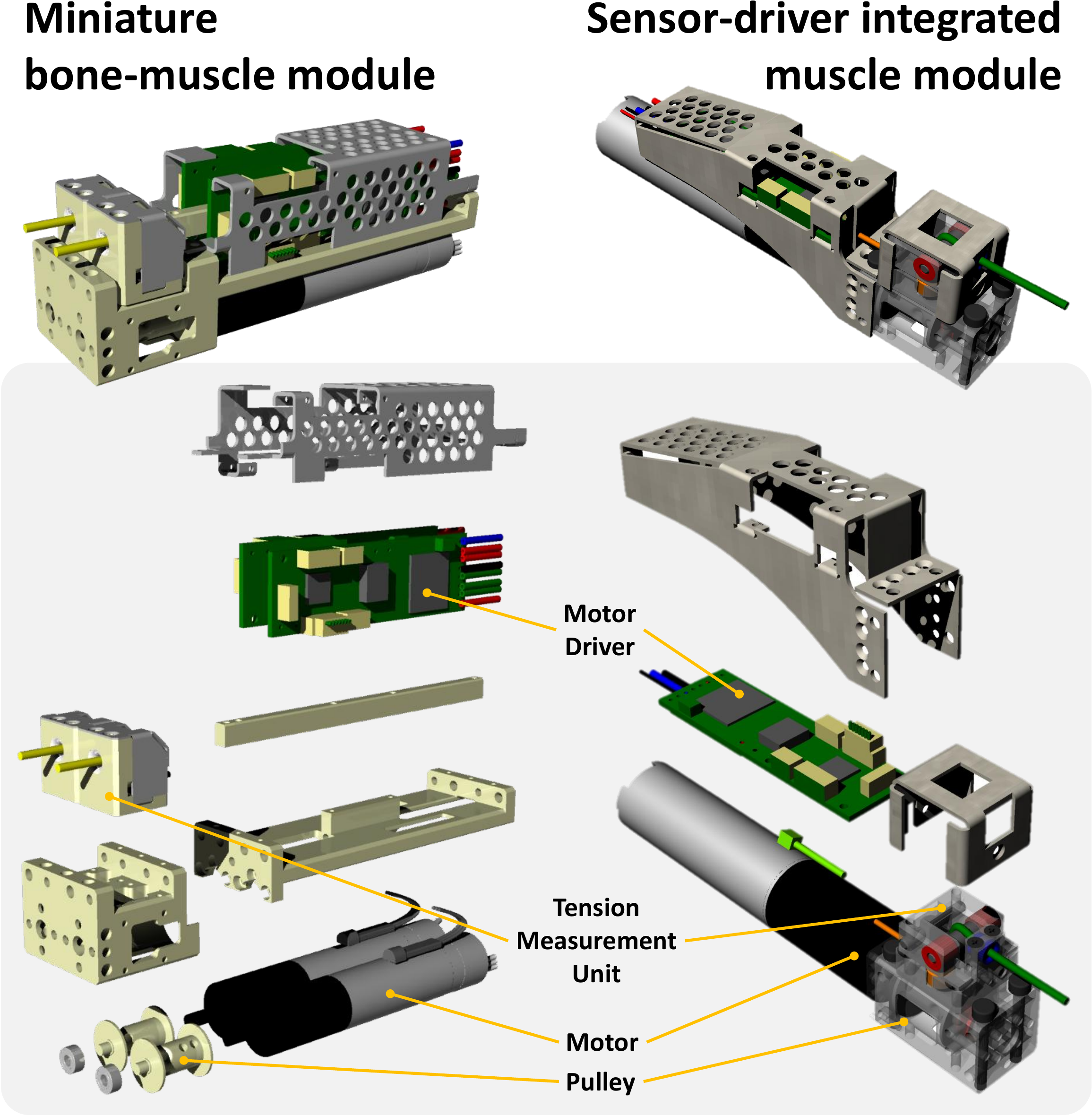}
  \vspace{-1.0ex}
  \caption{Conventional muscle modules: composed of an elongated motor with a high gear ratio and pulley, a tension measurement unit, a motor driver, and their protective covers.}
  \label{figure:comparison}
  \vspace{-3.0ex}
\end{figure}

\begin{table*}[t]
  \centering
  \caption{Comparison among miniature bone-muscle module \cite{kawaharazuka2017forearm}, sensor-driver integrated muscle module \cite{asano2015sensordriver}, and QDD-MM.}
  \vspace{-2.0ex}
  \scalebox{0.9}{
    \begin{tabular}{c|c|c|c}
      \hline
      & Miniature bone-muscle module \cite{kawaharazuka2017forearm} & Sensor-driver integrated muscle module \cite{asano2015sensordriver} & QDD-MM (this study) \\ \hline \hline
      Module dimensions [${\textrm{mm}}^{3}$]& $32.0\times40.0\times126$ & $22.0\times40.5\times149$ & $46.0\times58.0\times82$ \\ \hline
      Module weight [kgf] & 0.30 & 0.32 & 0.34 \\ \hline
      Number of actuators & 2 & 1 & 1\\ \hline
      Actuator type & Maxon BLDC-60W & Maxon BLDC-120W & RobStride05 \\ \hline
      Gear ratio & 84:1 & 53:1 & 7.75:1 \\ \hline
      Pulley diameter [mm] & 8 & 12 & 24 (12) \\ \hline
      Continuous maximum tension (motor) [N]\textsuperscript{*1} & 260 & 284 & 133 (267) \\ \hline
      Continuous maximum tension (gear) [N]\textsuperscript{*2} & 100 & 133 & - \\ \hline
      No-load winding speed [mm/s] & 216 & 200 & 603 (302) \\ \hline
    \end{tabular}
  }
  \label{table:comparison}
  \begin{flushleft}
  \footnotesize
    \textsuperscript{*1} Since the original papers \cite{asano2015sensordriver, kawaharazuka2017forearm} did not take gear transmission efficiency into account, the reported values differ from those in the previous works. \\
    \textsuperscript{*2} Maximum continuous tension determined by gear strength
  \end{flushleft}
  \vspace{-4.0ex}
\end{table*}

\subsection{Comparison with Existing Muscle Modules} \label{subsec:comparison}
\switchlanguage%
{%
  When comparing the proposed QDD-MM with previously developed muscle modules \cite{asano2015sensordriver, kawaharazuka2017forearm} shown in \figref{figure:comparison}, the prior modules have a slender cylindrical shape and consist of a BLDC motor, a custom motor driver, a pulley for wire winding, and a tension measurement unit.
  Because these motors have low torque density and cannot generate high torque on their own, they are combined with high gear-ratio transmissions.
  As a result, the modules exhibit low backdrivability and slow actuation speed.
  In addition, due to the low transmission efficiency of the gears, estimating muscle tension from motor current is difficult, which necessitates the use of a tension sensor mounted on the wire routing unit.
  In contrast, QDD-MM employs a quasi-direct-drive motor, eliminating the need for a muscle tension sensor and greatly simplifying the overall structure while achieving high backdrivability and fast actuation.
  The method of attaching the wire to the pulley also differs significantly.
  In previous muscle modules, a hole was drilled orthogonally through the pulley, and the wire was passed through and tied on the opposite side.
  However, this design causes the wire to bend sharply with a minimum bending radius of approximately 0.1 mm near the starting point, placing almost all the load on the knot.
  This greatly reduces wire durability.
  QDD-MM improves wire handling by routing the wire alternately through seven small pulleys, increasing the minimum bending radius to 3 mm and fixing the wire through friction instead of relying on a knot.
  This substantially enhances durability.
  The wire material has also been updated: earlier systems used 3 mm or 2 mm Dyneema, whereas QDD-MM employs 2 mm Zylon, which has a higher elastic modulus and is therefore less prone to elongation.
  Furthermore, the connection between the motor and the pulley has been revised.
  Previous modules relied on a D-cut shaft secured with a set screw because the motor shaft was thin.
  In QDD-MM, the motor and pulley are instead directly connected using a dowel pin, providing a more robust and reliable connection.
  These improvements in the wire attachment method to the pulleys and in the connection between the motor and the pulley became possible only after the form factor of the muscle module changed from a slender shape to a stubby one, and they represent important advancements over previous designs.

  \tabref{table:comparison} presents a numerical comparison between QDD-MM and previous muscle modules.
  Although QDD-MM is slightly larger in width and height, its depth is significantly shorter, and its overall mass is nearly the same as that of earlier modules.
  In terms of gear ratio, previous muscle modules employed high ratios such as 84:1 and 53:1, whereas QDD-MM uses a markedly lower ratio of 7.75:1 with a single-stage planetary gearbox.
  The pulley diameter of QDD-MM can be freely adjusted.
  For example, when set to 12 mm, the maximum continuous tension determined by motor limitations becomes 260 N, 284 N, and 267 N for each muscle module, enabling QDD-MM to achieve tension levels comparable to previous designs.
  Despite this, the wire winding speed is approximately 1.5 times faster than that of existing muscle modules, and the low gear ratio results in a substantial improvement in backdrivability.
  In addition, previous muscle modules suffered from insufficient gear strength; in practice, the gears would fail if forces of 100 N or 133 N were applied continuously.
  QDD-MM avoids this issue due to its low gear ratio.
  Note that, in this study, we prioritize fast actuation, and therefore the pulley diameter was set to 24 mm in our implementation.

  In this way, applying QDD motors to muscle modules resolves many of the longstanding challenges found in previous designs.
  The force and speed characteristics reflect both the selected QDD actuator and the pulley radius; they are not independent innovations of the module.
  Beyond the QDD cable actuation demonstrated in \cite{mi2025design,sawaguchi2024vlimb}, QDD-MM provides a reusable mechanical interface for changing muscle number, routing, and structural arrangement, and evaluates the resulting module against conventional muscle modules.
  Its open-source design complements existing open-source tendon-driven robots \cite{deutschmann2022open} by providing a muscle unit that can be reused across robot designs.
}%
{%
  ここでは, 本研究で提案したQDD-MMと, これまでに提案されてきた筋モジュールを比較する.
  これまでの筋モジュール\cite{asano2015sensordriver, kawaharazuka2017forearm}を\figref{figure:comparison}に示す.
  これまでの筋モジュールは細長い筒状の形状をしており, BLDCモータと自作のモータドライバ, ワイヤを巻き取るプーリ, 張力測定ユニットから構成されている.
  モータ自体が細長い形状をしており, 大きなトルクを発揮できないため, 高い減速比のギアを組み合わせている.
  そのため, バックドライバビリティが低く, 動作速度も遅い.
  ギアの伝達効率が低いため, 電流から筋張力を推定することが難しく, wire routing unitに張力センサを搭載した張力測定ユニットが必要となっている.
  これに対してQDD-MMは, 準ダイレクトドライブモータを用いることで, 筋張力センサを不要として構造を大幅に簡素化し, かつ高いバックドライバビリティと高速動作を実現している.
  プーリに対するワイヤの取り付け方法であるが, これまでの筋モジュールはプーリに直交して穴を開け, ここにワイヤを通し反対側で結んでいる.
  しかしこれでは, ワイヤが始点付近で曲げ半径0.1 mm程度で急激に曲がり, ほとんどの力をワイヤ始点の結び目だけで持つため, ワイヤの耐久性が低くなってしまう.
  QDD-MMでは, ワイヤを7つの小さなプーリに交互に通すことで, 最小曲げ半径を3 mmまで向上させ, かつ結び目ではなく摩擦によってワイヤを固定することで, その耐久性を大幅に向上させている.
  なお, ワイヤ自体の素材も変更しており, これまではDyneemaの3 mmまたは2 mmを用いていたが, QDD-MMではより高い弾性率(ワイヤが伸びにくい)を誇るZylonの2 mmを用いている.
  また, モータに対するプーリの接続も変更されており, これまではモータ軸が細いためDカットシャフトとイモネジにより接続していたが, QDD-MMではダウウェルピンを用いてモータとプーリを直接, より確実に接続できるようになっている.
  これらプーリに対するワイヤ取り付け方法やモータとプーリの接続方法は, これまでの筋モジュールのslenderな形がstubbyな形状へと変化したことで初めて可能となった重要な改良である.

  \tabref{table:comparison}に, QDD-MMとこれまでの筋モジュールの比較を数値で示す.
  まず, QDD-MMはこれまでの筋モジュールより幅と高さは上がっているが, 奥行きが大幅に短くなっており, モジュールの重さ自体はほとんど同じである.
  減速比を見ると, これまでの筋モジュールは84:1や53:1が用いられていたのに対して, QDD-MMは7.75と非常に低く, 一段の遊星減速機となっている.
  QDD-MMのプーリ直径は自由に変更できるが, 例えば12 mmにした場合, モータ自体の制約から来る最大連続張力は各筋モジュールについて260 N, 284 N, 267 Nとなり, QDD-MMはこれまでの筋モジュールと同等の張力を発揮できる.
  それにも関わらず, ワイヤ巻取り速度はこれまでの筋モジュールに比べ約1.5倍高速であり, かつ減速比が低いため, バックドライバビリティも大幅に向上している.
  加えて, これまでの筋モジュールはギアの強度が足りておらず, 実際には100 Nや133 N以上を出し続けるとギアが壊れてしまうという問題があった.
  一方で, QDD-MMは低減速比のためこの点も解消されている.
  なお, 本研究ではより高速動作を重視しているため, 実際にはプーリ直径を24 mmと設定して作成している.

  このように, QDDモータを筋モジュールに用いることで, これまで存在していた様々な課題を一挙に解決することができる.
  張力・速度特性は選定したQDDアクチュエータとプーリ半径に依存し, モジュール独自の新規性とは区別される.
  先行研究\cite{mi2025design,sawaguchi2024vlimb}で実現されているQDDケーブル駆動に対し, QDD-MMは筋数, ワイヤ経路, 構造配置を変更する共通の機械インタフェースを提供し, 従来の筋モジュールとの比較によって評価している.
  また, 既存のオープンソース腱駆動ロボット\cite{deutschmann2022open}に対し, ロボット設計をまたいで再利用できる筋単位の公開設計を提供する.
}%

\section{Experiments} \label{sec:experiment}

\subsection{Basic Performance Experiment} \label{subsec:basic-exp}
\begin{figure}[t]
  \centering
  \includegraphics[width=0.95\columnwidth]{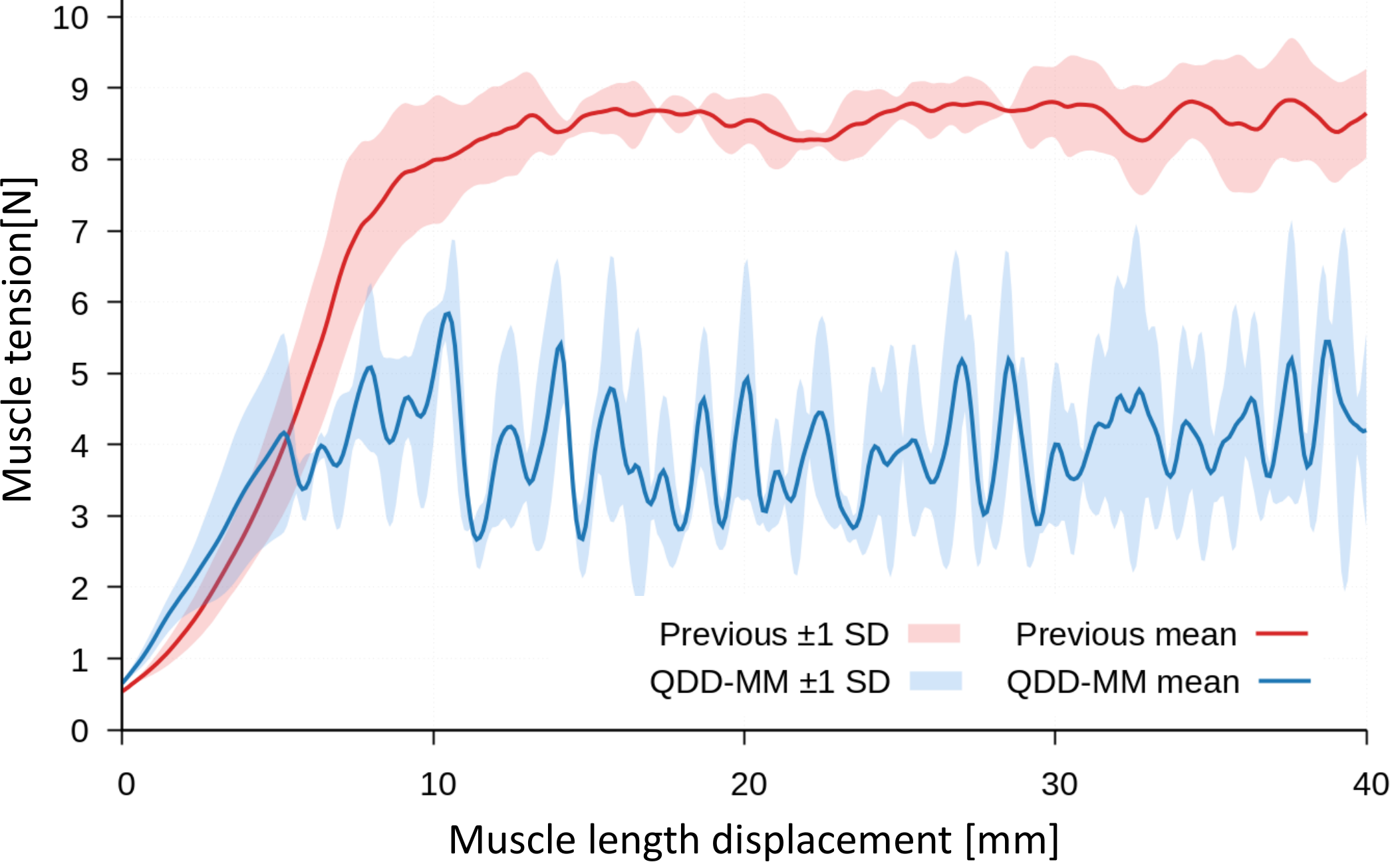}
  \vspace{-1.0ex}
  \caption{The relationship between the muscle length displacement and muscle tension for the previous muscle module \cite{asano2015sensordriver} (Previous) and QDD-MM.}
  \label{figure:exp-backdrivability}
  \vspace{-3.0ex}
\end{figure}

\begin{figure}[t]
  \centering
  \includegraphics[width=0.75\columnwidth]{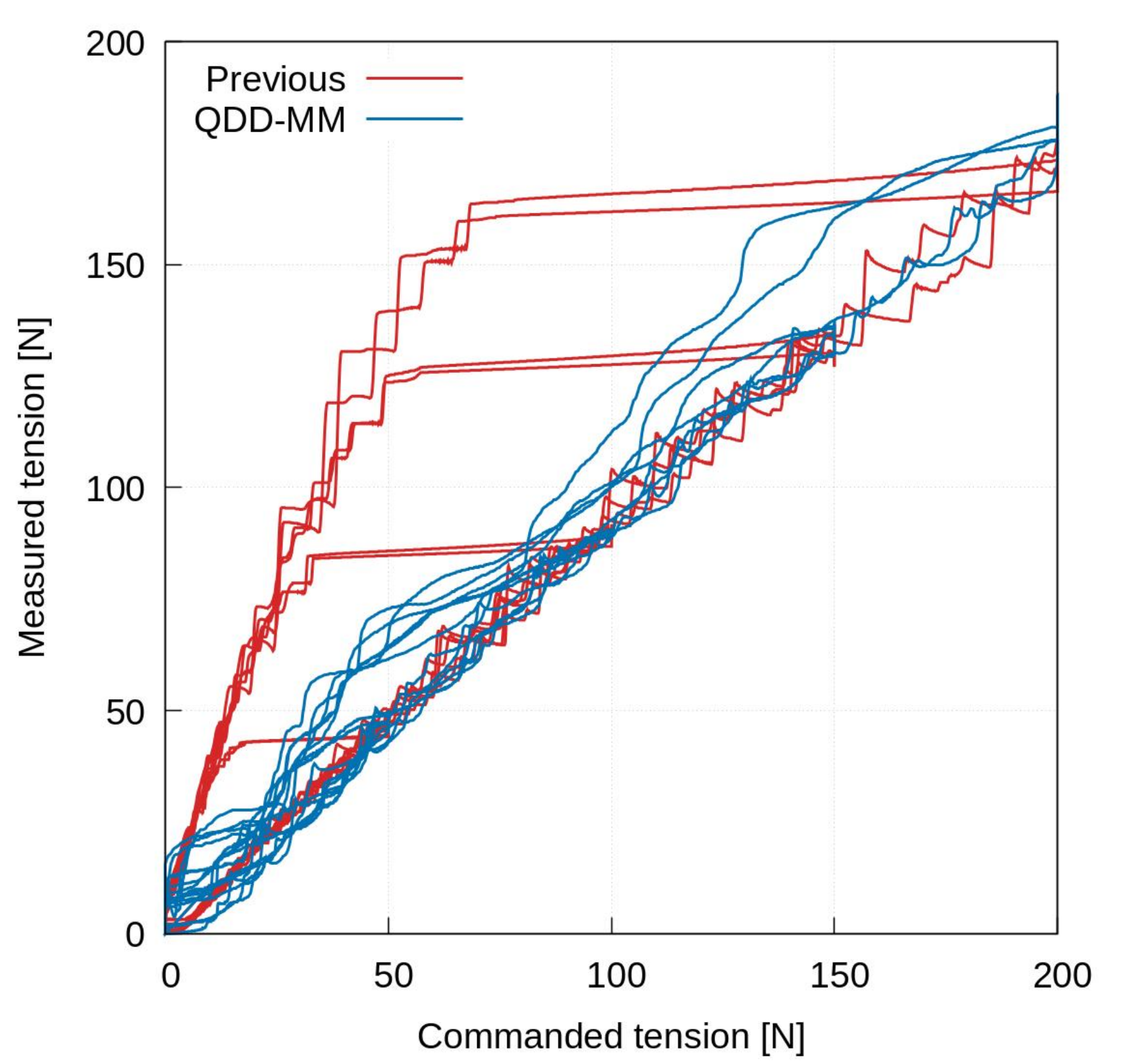}
  \vspace{-1.0ex}
  \caption{The relationship between the commanded and measured muscle tensions for the previous muscle module \cite{asano2015sensordriver} (Previous) and QDD-MM.}
  \label{figure:exp-hysteresis}
  \vspace{-3.0ex}
\end{figure}

\switchlanguage%
{%
  We conducted experiments to evaluate the basic performance of the proposed QDD-MM and a previous muscle module (Previous) \cite{asano2015sensordriver}.
  The module in \cite{kawaharazuka2017forearm} was not included in the comparison, as its specifications are nearly identical to those in \cite{asano2015sensordriver}.

  First, we conducted an experiment to evaluate backdrivability.
  The QDD-MM and the previous muscle module were mounted on a tabletop tensile testing setup, and the wire was pulled at a constant speed of 5 mm/s.
  Backdrivability was then evaluated from the resulting resistive force.
  The relationship between wire extension and resistive force was measured three times for each module, and the mean and variance were calculated.
  The results are shown in \figref{figure:exp-backdrivability}.
  While the previous muscle module exhibited a resistive force of approximately 8 N, the QDD-MM showed a resistive force of only about 4 N, indicating that the QDD-MM has approximately twice the backdrivability of the previous module.
  On the other hand, because the QDD-MM employs a quasi-direct-drive motor, it exhibits relatively large cogging torque, resulting in larger fluctuations in the resistive force.

  Next, we conducted an experiment to investigate the relationship between the commanded muscle tension and the actual muscle tension.
  Each muscle module and a force gauge were aligned in a straight line.
  Current control was applied to increase the commanded tension from 0 to $f^{ref}$ over 10 seconds, followed by a 5-second hold, and then decreased back to 0 over another 10 seconds.
  The value of $f^{ref}$ was set to 50, 100, 150, and 200 N, and each condition was repeated twice.
  We then measured the relationship between the commanded muscle tension and the actual tension obtained from the force gauge.
  The conversion from commanded tension to motor current considered the pulley radius, gear ratio, and gear transmission efficiency, although losses at the Top-Pulley, Motor-Pulley, and the tension measurement unit were not included.
  The results are shown in \figref{figure:exp-hysteresis}.
  Both muscle modules achieved approximately 90\% of the commanded tension, and we attribute the remaining 10\% loss to friction within the pulleys and tension measurement unit, which were not accounted for in the conversion model.
  Here, hysteresis means that the measured tension at the same command depends on whether the command is increasing or decreasing.
  Each cycle follows an increasing branch from 0 to $f^{ref}$, a hold at $f^{ref}$, and a decreasing branch back to 0.
  The vertical separation between the two branches at the same command indicates the history dependence.
  The most significant difference between the previous muscle module and QDD-MM is found in these hysteresis characteristics.
  Due to the high gear ratio and low backdrivability of the previous module, substantial hysteresis occurs between the pulling and releasing phases.
  For example, when the command tension is reduced from 200 N to 70 N, the previous module fails to unwind and still produces about 160 N of tension.
  In contrast, QDD-MM produces only about 80 N under the same condition, demonstrating considerably smaller hysteresis.
  In other words, while it was almost impossible to estimate muscle tension from motor current in previous modules, QDD-MM exhibits an approximately proportional relationship between motor current and actual muscle tension.
  This supports current-based tension estimation without a dedicated tension sensor in the demonstrated applications, while residual friction and hysteresis still introduce estimation error.
}%
{%
  これまでの筋モジュール(Previous) \cite{asano2015sensordriver} (\cite{kawaharazuka2017forearm}は\cite{asano2015sensordriver}とほとんど同じ設定のため, 近い特性を持っている)とQDD-MMの基本性能を評価する実験を行った.

  まず, バックドライバビリティを測定する実験を行った.
  卓上引張試験にQDD-MMとこれまでの筋モジュールを取り付け, ワイヤを一定速度(5 mm/s)で引っ張ることで, その際の抵抗力からバックドライバビリティを評価した.
  ワイヤの伸展量に対する抵抗力の推移をそれぞれ3回ずつ計測し, その平均と分散を求めた.
  その結果を\figref{figure:exp-backdrivability}に示す.
  これまでの筋モジュールは約8 Nの抵抗力が生じるのに対し, QDD-MMは約4 Nと, 2倍程度のバックドライバビリティを有していることがわかる.
  一方で, QDD-MMは準ダイレクトドライブモータを用いているためコギングトルクが大きく, 抵抗力の変動が大きいことがわかる.

  次に, 指令筋張力と実際の筋張力の関係を計測する実験を行った.
  各筋モジュールとフォースゲージを直線上に並べ, 指令張力が0から$\bm{f}^{ref}$になるように10秒かけて電流を印加 (電流制御を用いる), その後5秒待ち, 10秒かけて戻した.
  $\bm{f}^{ref}$をそれぞれ順に50, 100, 150, 200 Nに変化させることを2回ずつ繰り返し, その際の指令筋張力とフォースゲージから得られる実際の筋張力の関係を計測した.
  なお, 指令筋張力から電流への変換は, プーリ半径, ギア比, ギアの伝達効率を考慮して行っているが, Top-PulleyやMotor-Pulley, Tension Measurement Unitにおける損失は考慮していない.
  その結果を\figref{figure:exp-hysteresis}に示す.
  どちらの筋モジュールも, 指令値に対して, 90\%程度の筋張力が実際に発揮できていることがわかる.
  この損失10\%は, 考慮できていない各種プーリや張力測定ユニットにおける摩擦損失によるものであると考えられる.
  ここでヒステリシスとは, 同一の指令張力であっても, 指令の増加・減少方向と直前の最大値$f^{ref}$によって実測張力が異なることである.
  各試行は0から$f^{ref}$への増加, 最大値での保持, 0への減少の順に進む. 同一指令値における増加時と減少時の軌跡の縦方向の差が履歴依存性を表す.
  これまでの筋モジュールとQDD-MMの最大の差は, このヒステリシスにある.
  これまでの筋モジュールはギア比が高く, バックドライバビリティが低いため, ワイヤを引っ張る時と戻す時で大きなヒステリシスが発生している.
  例えば, これまでの筋モジュールは指令張力を200 Nから70 Nに下げてもモータが巻き戻らず160 Nの力を発揮してしまうのに対して, QDD-MMは80 Nと, ヒステリシスが非常に小さいことがわかる.
  つまり, これまでのモジュールでは電流から筋張力を推定することはほとんど不可能であったが, QDD-MMでは電流と実際の筋張力がほぼ比例するため, 本実験の用途では専用張力センサを用いない電流に基づく張力推定が可能であることが示された. ただし, 残留摩擦とヒステリシスによる推定誤差は存在する.
}%

\begin{figure*}[t]
  \centering
  \includegraphics[width=1.8\columnwidth]{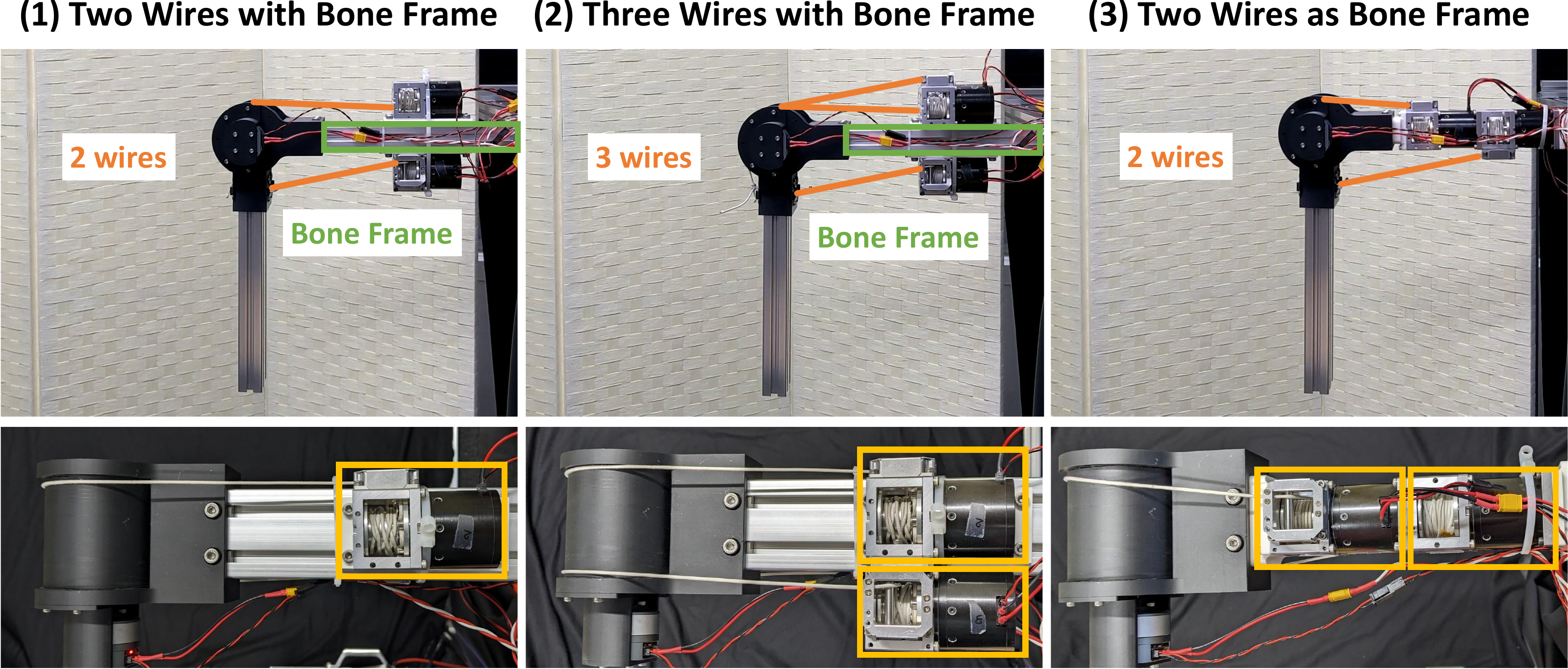}
  \vspace{-1.0ex}
  \caption{Experimental setup: the left panel shows a configuration with two muscles attached to the bone frame, the middle panel shows a configuration with three muscles realized by adding a module in parallel, and the right panel shows a series connection in which the module itself is used as the bone frame.}
  \label{figure:exp-setup}
  \vspace{-1.0ex}
\end{figure*}

\begin{figure*}[t]
  \centering
  \includegraphics[width=1.9\columnwidth]{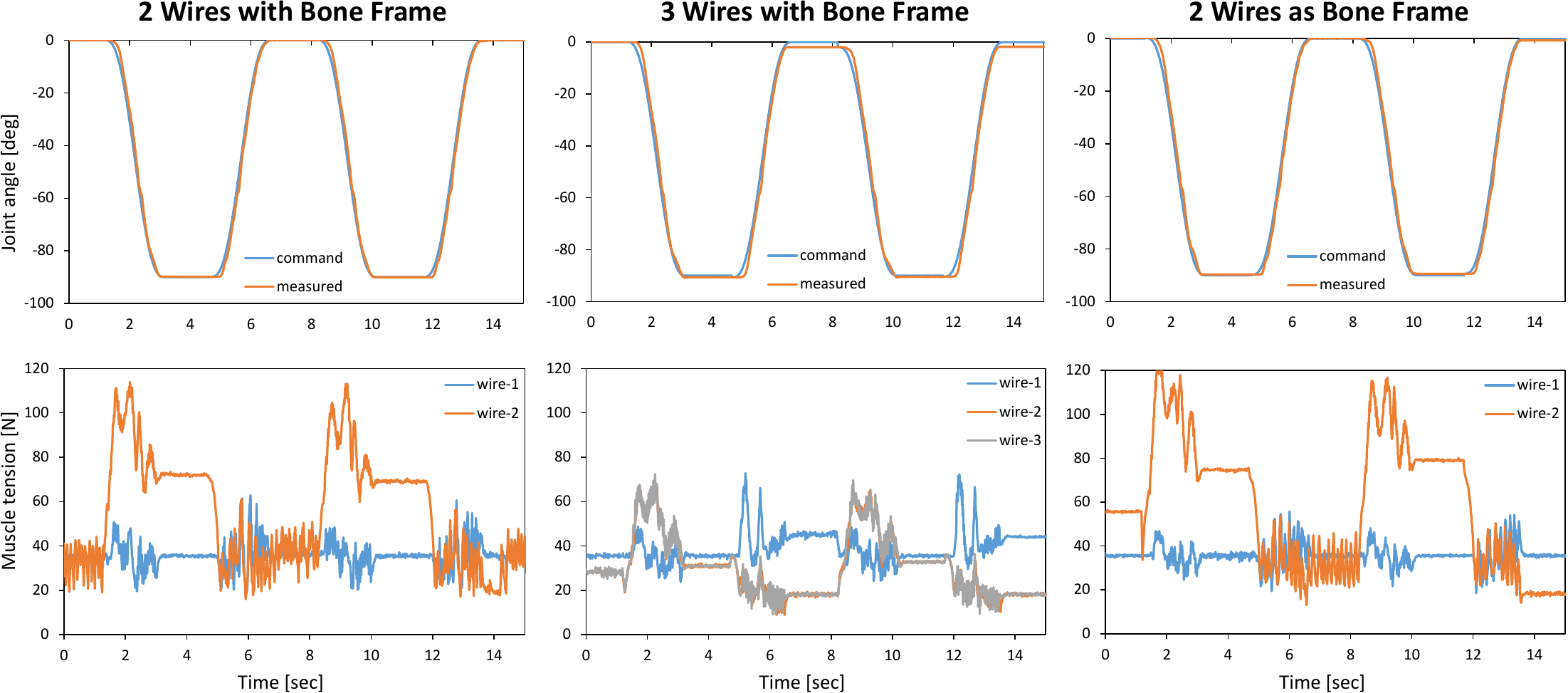}
  \vspace{-1.0ex}
  \caption{The experimental results corresponding to \figref{figure:exp-setup}, including the commanded and measured joint angles as well as the measured muscle tensions.}
  \label{figure:exp-result}
  \vspace{-3.0ex}
\end{figure*}

\subsection{1-DOF Tendon-driven Robot Experiment} \label{subsec:exp-result}
\switchlanguage%
{%
  This experiment evaluates reconfigurability, joint-angle tracking, and load sharing using one-joint robots with two or three QDD-MMs.
  \figref{figure:exp-setup} illustrates examples of the robot configurations used in the experiments.
  Configuration (1) is a one-joint--two-muscle tendon-driven robot, in which two QDD-MM units are attached to a bone frame using module connectors.
  Configuration (2) extends (1) by using an additional muscle module in parallel via a module connector, thereby creating a one-joint--three-muscle robot.
  Configuration (3) connects two muscle modules in series via the module connector so that the modules themselves function as the skeletal structure, resulting in a compact one-joint--two-muscle tendon-driven robot.

  For each of the three robots, the commanded joint angle was moved from 0 to -90 degrees over 2 seconds, held for 1.5 seconds, and then returned over another 2 seconds.
  This cycle was repeated while measuring joint angles and muscle tensions.
  The control method was a standard muscle-tension-based approach: the joint-angle error was converted into joint torque using PD control, and the required torque was allocated to the muscles \cite{jantsch2012computed}.
  The minimum tension for the antagonist muscle (wire-1) was set to 30 N, and the minimum tension for the agonist muscles (wire-2 or wire-3) was set to 15 N.
  The results are shown in \figref{figure:exp-result}.
  All three tendon-driven robots were able to track the commanded joint angles with reasonably good accuracy.
  Regarding muscle tension, configurations (1) and (3), which both use two muscles, exhibited similar tension profiles.
  In contrast, configuration (2), which uses three muscles, demonstrated cooperative behavior between wire-2 and wire-3, reducing the required tension for each muscle by approximately half and significantly lowering the load.
  These results demonstrate that QDD-MM enables compact tendon-driven robots by allowing additional muscles to be added through modular connections and by using the muscle modules themselves as structural components.
}%
{%
  ここでは, QDD-MMを用いた1関節2筋または1関節3筋の腱駆動ロボットを制作し, 再構成可能性, 関節角度追従, 筋間の負荷分担を評価する.
  \figref{figure:exp-setup}に, 実験で用いた腱駆動ロボットの構成例を示す.
  (1)は, 1関節2筋の腱駆動ロボットであり, QDD-MMを2つ用いて, それらをmodule connectorでbone frameに取り付けている.
  (2)は, (1)に対してmodule connectorで筋モジュールを並列に追加することで, 1関節3筋の腱駆動ロボットを実現している.
  (3)は, module connectorで筋モジュール同士を直列に接続し, 筋モジュール自体を骨格として利用することで, コンパクトに1関節2筋の腱駆動ロボットを実現している.

  これら3種類の腱駆動ロボットに対して, それぞれ指令関節角度を0から-90度まで2秒かけて動かし, 1.5秒止まり, また2秒かけて戻る動作を繰り返し行い, その際の関節角度と筋張力を計測した.
  制御は一般的な筋張力制御を用いており, 関節角度の誤差をPD制御により関節トルクに変換し, これを筋張力に分配している\cite{jantsch2012computed}.
  なお, 拮抗筋であるwire-1の最低張力を30 N, 主動筋であるwire-2またはwire-3の最低張力を15 Nに設定している.
  その結果を\figref{figure:exp-result}に示す.
  3つの腱駆動ロボット全てにおいて, 指令関節角度に対して比較的良好に追従できていることがわかる.
  筋張力について, 2筋である(1)と(3)は両者とも同じような筋張力遷移を示している.
  これに対して, 3筋である(2)では, 主動筋であるwire-2とwire-3が協調して動作しており, 必要な筋張力が半分になることで, 負荷が大きく軽減されていることがわかる.
  このように, QDD-MMを用いることで, 追加の筋肉をモジュール間接続により増やしたり, 筋モジュール自体を骨格として利用したりすることで, コンパクトに腱駆動ロボットを実現できることが示された.
}%

\begin{figure*}[t]
  \centering
  \includegraphics[width=1.9\columnwidth]{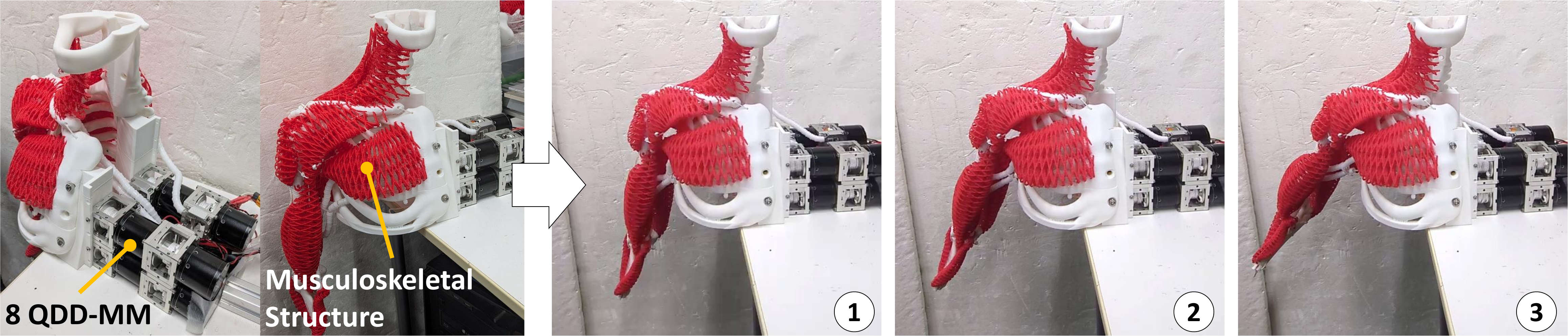}
  \vspace{-1.0ex}
  \caption{Shoulder--arm prototype driven by eight externally mounted QDD-MMs. Left: module arrangement and musculoskeletal structure. Right, (1)--(3): successive postures showing the change in upper-arm orientation and shoulder configuration relative to the torso.}
  \label{figure:muscle-exp}
  \vspace{-3.0ex}
\end{figure*}

\subsection{Musculoskeletal Robot Experiment} \label{subsec:muscle-exp}
\switchlanguage%
{%
  Here, we construct the arm of a musculoskeletal humanoid using eight QDD-MMs and a 3D-printed musculoskeletal structure~\cite{yoshimura2024psm}.
  The purpose is to demonstrate integration of multiple identical modules with anatomical wire routing to actuate the scapula and upper arm together.
  The overall system and its motions are shown in \figref{figure:muscle-exp}.
  The eight QDD-MMs are connected both vertically and horizontally, resulting in a compact and space-efficient arrangement.
  By actuating these modules, we successfully achieve simultaneous movements of the scapula and the upper arm.
  The numbered images (1)--(3) show successive postures: the upper arm moves from a near-downward posture toward a more outward posture while the shoulder configuration changes.
  Although the motor units are currently located outside the body, our future plan is to integrate them within the body and construct a full musculoskeletal humanoid.
}%
{%
  ここでは, 筋骨格ヒューマノイドの腕部を, 8つのQDD-MMと3Dプリンタによる筋骨格構造の造形\cite{yoshimura2024psm}により構築する.
  本実験の目的は, 同一の筋モジュールを解剖学的なワイヤ配置と組み合わせ, 肩甲骨と上腕を同時に駆動できることを示すことである.
  その全体像と動きを\figref{figure:muscle-exp}に示す.
  8つのQDD-MMは縦・横に接続され, 無駄なくコンパクトに収まっている.
  これらをアクチュエートすることで, 肩甲骨と上腕を同時に動かすことに成功している.
  連続姿勢(1)--(3)では, 上腕が下向きに近い姿勢から外側へ移動するとともに肩周辺の姿勢が変化する.
  現在はモータ部が外に出ている状態であるが, 今後これを体内に納め, 筋骨格ヒューマノイド全体を構築する予定である.
}%

\section{CONCLUSION} \label{sec:conclusion}
\switchlanguage%
{%
  In this study, we proposed QDD-MM, a simple and compact muscle module that utilizes quasi-direct-drive motors with the aim of further democratizing tendon-driven robots.
  By leveraging recent advances in quasi-direct-drive motors, we were able to eliminate the need for tension sensors and greatly simplify the structure of conventional muscle modules, which rely on high gear ratios and dedicated sensing units.
  At the same time, QDD-MM achieves high backdrivability and fast actuation.
  As a result, hysteresis is significantly reduced, and muscle modules can be connected in parallel or in series, allowing the number of muscles to be freely increased and enabling the modules themselves to function as part of the robot's skeletal structure.
  These findings suggest that the characteristics of modern quasi-direct-drive motors are highly compatible with tendon-driven robotic systems.
  In future work, we plan to develop more complex tendon-driven robots using QDD-MM and to investigate optimization methods for muscle arrangement.
  We also intend to evaluate the durability of QDD-MM and to examine its maintainability under long-term operation.
}%
{%
  本研究では, 今後の腱駆動ロボットをより民主化することを目的として, 準ダイレクトドライブモータを活用したシンプルでコンパクトな筋モジュールQDD-MMを提案した.
  これまでの高減速比かつ張力センサを必要とする筋モジュールを, 現在の準ダイレクトドライブモータの発展を活用することで, 筋張力センサを不要として構造を大幅に簡素化し, かつ高いバックドライバビリティと高速動作を実現した.
  結果として, ヒステリシスが大幅に下がるだけでなく, モジュールを並列や直列に接続することで, 自由に筋数を増やしたり, 筋モジュール自体を骨格として利用することが可能となった.
  現在の準ダイレクトモータの特性が, 腱駆動ロボットに非常にマッチしていることが示されたと言える.
  今後, このQDD-MMを用いたより複雑な腱駆動ロボットの制作や, 筋配置の最適化手法の開発などを進めていく予定である.
  また, QDD-MMの耐久性評価や, 長期的な使用におけるメンテナンス性の検証も今後行っていきたい.
}%

\section*{ACKNOWLEDGMENT}
This work was partially supported by JST Moonshot R\&D Program Japan Grant Number JPMJMS263D.

{
  \bibliographystyle{IEEEtran}
  \bibliography{main}
}

\end{document}